\documentclass[conference]{IEEEtran}

\IEEEoverridecommandlockouts
\usepackage{cite}
\usepackage{amsmath,amssymb,amsfonts}
\usepackage{multirow}
\usepackage{graphicx}
\usepackage{subfig}
\usepackage{textcomp}
\usepackage{xcolor}
\usepackage[utf8]{inputenc}
\usepackage[english]{babel}
\usepackage{fancyhdr}
\usepackage{array}
\usepackage{tabularx}
\usepackage[ruled,vlined]{algorithm2e}
\usepackage{algpseudocode}
\usepackage[section]{placeins}
\usepackage{float}

\usepackage{tikz}
\usetikzlibrary{quantikz}
\usepackage{amsmath}     
\usepackage{amssymb}     
\usepackage{physics}     
\usepackage{graphicx}

\tikzset{meter/.append style={draw, inner sep=8, rectangle, font=\vphantom{A}, minimum width=30, line width=.8,
 path picture={\draw[black] ([shift={(.1,.3)}]path picture bounding box.south west) to[bend left=50] ([shift={(-.1,.3)}]path picture bounding box.south east);\draw[black,-latex] ([shift={(0,.1)}]path picture bounding box.south) -- ([shift={(.3,-.1)}]path picture bounding box.north);}}}

\usepackage{amsthm}

\def\BibTeX{{\rm B\kern-.05em{\sc i\kern-.025em b}\kern-.08em
    T\kern-.1667em\lower.7ex\hbox{E}\kern-.125emX}}

\begin{document}

\title{CVaR-Based Variational Quantum Optimization \\ for User Association in Handoff-Aware \\ Vehicular Networks     
}

\author{
\IEEEauthorblockN{
    Zijiang~Yan\IEEEauthorrefmark{1}, 
    Hao~Zhou\IEEEauthorrefmark{2},
    Jianhua~Pei\IEEEauthorrefmark{1}\IEEEauthorrefmark{3}, 
    Aryan~Kaushik\IEEEauthorrefmark{4},
    Hina~Tabassum\IEEEauthorrefmark{1} and
    Ping~Wang\IEEEauthorrefmark{1}
}
\IEEEauthorblockA{\IEEEauthorrefmark{1} Department of Electrical Engineering and Computer Science, York University, Toronto, ON, Canada }
  \IEEEauthorblockA{\IEEEauthorrefmark{2} Department of Computer Science,  McGill University, Montréal, QC, Canada}
  \IEEEauthorblockA{\IEEEauthorrefmark{3} School of Electrical and Electronic Engineering,  Huazhong University of Science and Technology, Wuhan, China}
  \IEEEauthorblockA{\IEEEauthorrefmark{4} Department of Computing and Mathematics,  Manchester Metropolitan University, Manchester, UK}
  \thanks{
  Authors e-mails are zjiyan@yorku.ca;   hao.zhou4@mail.mcgill.ca;     jianhuapei@hust.edu.cn;   a.kaushik@ieee.org; hinat@yorku.ca; pingw@yorku.ca  }
}

\raggedbottom

\maketitle
\begin{abstract}

Efficient resource allocation is essential for optimizing various tasks in wireless networks, which are usually formulated as generalized assignment problems (GAP). GAP, as a generalized version of the linear sum assignment problem, involves both equality and inequality constraints that add computational challenges. In this work, we present a novel Conditional Value at Risk (CVaR)-based Variational Quantum Eigensolver (VQE) framework to address GAP in vehicular networks (VNets). Our approach leverages a hybrid quantum-classical structure, integrating a tailored cost function that balances both objective and constraint-specific penalties to improve solution quality and stability. Using the CVaR-VQE model, we handle the GAP efficiently by focusing optimization on the lower tail of the solution space, enhancing both convergence and resilience on noisy intermediate-scale quantum (NISQ) devices. We apply this framework to a user-association problem in VNets, where our method achieves 23.5\% improvement compared to the deep neural network (DNN) approach.  
\end{abstract}

\begin{IEEEkeywords}
quantum machine learning, variation quantum eigensolver (VQE), generalized assignment problem (GAP), multiband VNet, Handoffs
\end{IEEEkeywords}

\section{Introduction}
\thispagestyle{fancy}

Resource allocation is a fundamental and crucial problem in wireless networks, which can be usually formulated as generalized assignment problems (GAP) and known as NP-hard ~\cite{abc}.
Allocation problems appear in various contexts, including computation offloading in edge computing systems, sub-channel assignment\cite{yan2022gc}, user scheduling and load balancing~\cite{tabassum2015spectral}, and antenna selection \cite{asaad2024joint}.
Traditional approaches to solving GAP in wireless networks often rely on classical optimization methods, which become computationally prohibitive as network complexity increases, particularly in emerging 5G/6G networks where channel coherence time is significantly shorter at higher frequencies such as mm-wave and THz bands. 
Moreover, machine learning techniques have been introduced to optimize the GAP problem through various approaches, including deep reinforcement learning (DRL) for vehicular communication optimization \cite{yan2022gc,yan2024Optimizing,yan2024generalized}, deep neural network (DNN)-based optimization for user-to-base station (BS) associations \cite{kaushik2021deep}, and over-the-air federated learning for antenna selection in massive MIMO systems \cite{asaad2024joint}.

The emergence of quantum computing has opened new frontiers in solving complex optimization problems across various domains.
Compared to classical machine learning approaches, the quantum algorithm operates within the quantum computational paradigm, offering potential advantages in accuracy and efficiency by directly leveraging quantum mechanical principles.
The potential advantages of quantum algorithms have been demonstrated in many existing studies \cite{abbas2021power,tilly2022variational} .
Among quantum algorithms, variational quantum eigensolver (VQE) \cite{tilly2022variational} has gained significant attention due to its hybrid quantum-classical nature, making it particularly suitable for implementation on noisy intermediate-scale quantum (NISQ) devices \cite{barkoutsos2020improving}.
While VQE was initially designed for quantum chemistry applications, its potential for solving optimization problems has sparked interest across different fields.

The application of VQE to the GAP problem in vehicular networks (VNet) presents a promising approach for several reasons.
\textbf{(1)} VQE's hybrid nature enables the exploitation of quantum advantages while mitigating hardware limitations through classical post-processing. Specifically, the quantum processor handles quantum state preparation, while the classical VQE network optimizes the parameters \cite{tilly2022variational}. In this work, we consider vehicle-to-BS associations are binary, making them suitable for mapping to a quadratic unconstrained binary optimization (QUBO) problem that aligns well with the quantum operators in VQE's framework.
\textbf{(2)} Compared to purely quantum algorithms, VQE demonstrates greater resilience to noise. Its variational approach allows the quantum system to achieve a low-energy state without requiring perfect operations, making it practical for NISQ devices \cite{sharma2020noise}.
\textbf{(3)} VQE employs variational quantum circuits with relatively shallow depths, which reduces the number of quantum gates required.\cite{fedorov2022vqe}. This design minimizes decoherence, making VQE feasible on current quantum hardware. Moreover, VQE allows for flexible ansatz customization, enabling adjustments across various vehicle-to-BS association scenarios.

 Motivated by the advantages of VQE techniques, this work considers a vehicular network scenario, aiming to optimize autonomous vehicle multi-band VNet associations (vehicle-to-BS). We propose a novel Conditional Value at Risk (CVaR)-based VQE approach to address vehicle-to-BS association problems, offering a robust solution to address uncertainties such as channel fading and handoff loss in vehicular networks. 
CVaR-VQE enhances the standard VQE by using a cost function that improves convergence and robustness, especially in NISQ devices. {This is achieved by focusing on the lowest energy of the quantum chemical system, thereby providing a better optimization approach.}
In particular, we minimize its CVaR value considering only the tail of the probability distribution to boost training efficiency. 
The integration of CVaR into the VQE framework facilitates a more conservative optimization strategy\cite{barkoutsos2020improving}, which is crucial for maintaining quality of service (QoS) in wireless communications.

The core contribution of this work is that we propose a novel CVaR-based variational quantum optimization technique for user association problems in vehicular networks.
Specifically, we model the vehicle-to-BS association as input quantum states, mapping wireless network constraints onto quantum circuits. Then we explore different ansatz configurations to balance accuracy and training efficiency in the VQE optimization process.
We integrate CVaR metrics into the quantum optimization framework to enhance solution robustness against network stochasticity.
Our approach demonstrates that the CVaR-based VQE framework can improve the average reward by 23.5\% with significantly reduced computational complexity compared to DNN methods.

\begin{figure}
    \centering
    \includegraphics[width=1\linewidth]{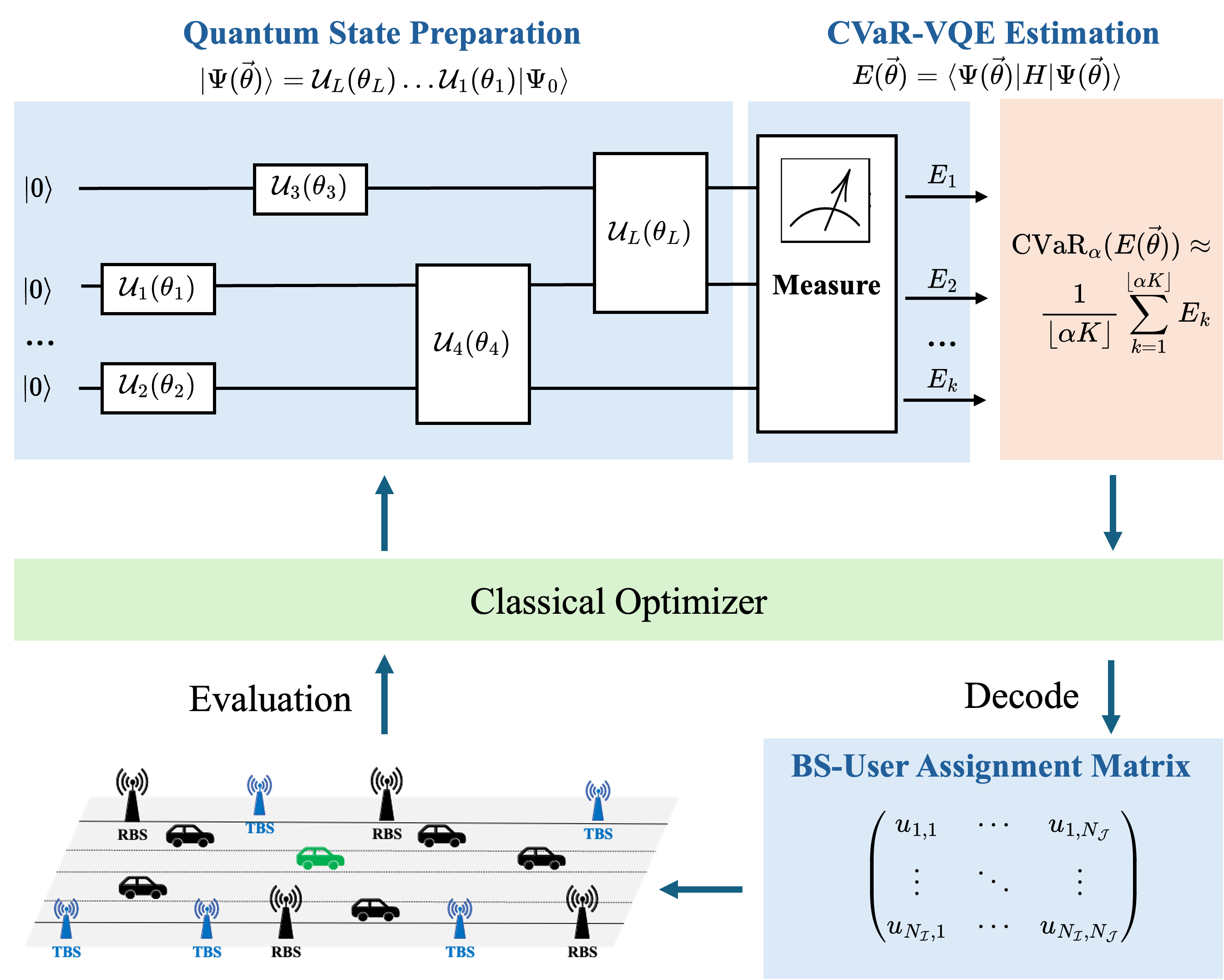}
    \caption{CVaR-VQE Framework with proposed VNet Model}
    \label{fig:illustration}
\end{figure}

\section{Vehicular Network System Model}
\label{sec:system_model}
This section presents the vehicular network model and the problem formulations employed in this study. Illustrated in Figure \ref{fig:illustration}, the model considers a downlink network environment comprising $N_R$ RF base stations (RBSs) and $N_T$ THz base stations (TBSs) interacting with multiple autonomous vehicles. We denote the set of autonomous vehicle users by $\mathcal{I} = \{1, \dots, I\}$ and base stations by $\mathcal{J} = \{1, \dots, J\}$.
Our goal, from a Vehicle-to-Infrastructure (V2I) communication perspective, is twofold: to maximize the V2I data rate and to minimize handovers.

We assume that $M_1$ autonomous vehicles (AVs) receive data from roadside BSs \cite{yan2022gc}, with each AV linked to a single BS, either RBS or TBS. These AVs use onboard units to collect real-time data from the vehicular network, including the speed, acceleration, and lane position of nearby vehicles.

The Signal-to-interference-plus-noise ratio (SINR) for the $j$-th AV from BS $i$ is given by \cite{yan2022gc}:

\begin{equation}{\label{sinr_rf}}
    \mathrm{SINR}^{\mathrm{RF}}_{ij} = \frac{P_{R}^{\mathrm{tx}}\:G_{R}^{\mathrm{tx}}\:G_{R}^{\mathrm{rx}} \left(\frac{c}{4\pi f_{R}} \right)^2 H_{i} }{r_{ij}^{\rho}\left(\sigma^2 + I_{R_j}\right)},
\end{equation}
where $P_{R}^{\mathrm{tx}}$, $G_{R}^{\mathrm{tx}}$, $G_{R}^{\mathrm{rx}}$, $c$, $f_{R}$, and $\rho$ represent the RBS transmit power, transmitting and receiving antenna gains, the speed of light, RF carrier frequency, and the path-loss exponent, respectively. Here, $r_{ij} = {(d_{ij}^2+h_{ij}^2)}^{1/2}$, where $d_{ij}$ is the 2D distance between AV $j$ and BS $i$, and $h_{ij}$ is the antenna height. Also, $H_i$ is the channel gain from RBS $i$, $\sigma^2$ is the receiver noise power, and $I_{R_j}$ is the cumulative interference from other RBSs. For TBSs, the SINR for AV $j$ is modeled as \cite{mobility}:

\begin{equation}
    \mathrm{SINR}^{\mathrm{THz}}_{ij} = \frac{G_{T}^{\mathrm{tx}} G_{T}^{\mathrm{rx}} \left(\frac{c}{4\pi f_{T}} \right)^2 P_{T}^{\mathrm{tx}}\: \mathrm{exp}(-K_a(f_T) r_{ij})r_{ij}^{-2} }{N_{T_j} + I_{T_j}},
\end{equation}
where $G_{T}^{\text{tx}}$, $G_{T}^{\text{rx}}$, $P_{T}^{\text{tx}}$, $f_{T}$, $K_{a}(f_T), N_{T_j}$ and $ I_{T_j}$ represent the TBS transmit and receive antenna gains, transmit power, THz carrier frequency, {the medium's molecular absorption coefficient, the cumulative interference between AV $j$ and the interfering TBS $i$, molecular and thermal cumulative absorption noise, respectively\cite{yan2024generalized}.}

Each AV, equipped with a single antenna, relies on beam alignment to ensure its receiving beam aligns with the TBS's transmitting beam. The likelihood of main lobe alignment for both the user and interfering TBS is denoted by probability $q$, with interference calculated accordingly \cite{mobility}. The available bandwidths for RBS and TBS are $W_R$ and $W_T$, respectively, and the AV-BS link data rate is given by:

\begin{equation}
    R_{ij} = W_j \log_2(1 + \mathrm{SINR}_{ij}).
\end{equation}

Each RBS and TBS can support a maximum of $Q_R$ and $Q_T$ AVs, respectively \cite{yan2024generalized}. For a connection to be acceptable, SINR must exceed a threshold $\gamma_{th}$. As AVs travel, they switch to different BSs when $\mathrm{SINR}_{ij}$ drops below $\gamma_{th}$. Frequent handovers (HOs) lead to reduced data rates due to latency, so we introduce a HO penalty $\mu$, with TBSs incurring a higher penalty than RBSs. The weighted data rate between BS $i$ and AV $j$ is defined as:

\begin{equation}
    \text{WR}_{ij} = \frac{R_{ij}}{\min \left(Q_i, n_i \right)} (1 - \mu),
\end{equation}
where $Q_i$ corresponds to the BS type ($Q_R$ or $Q_T$), and $n_i$ is the number of active users on BS $i$.

The Generalized Assignment Problem (GAP) is formulated to optimize the allocation between BSs and AVs as follows:

\begin{equation}
\label{eq:optimization_problem}
\begin{aligned}
\mathcal{P}: \quad & \max_{u} \quad Z \triangleq \sum_{i=1}^{I} \sum_{j=1}^{J} u_{ij} \text{WR}_{ij}, \\
\text{s. t. \quad}&   \mathrm{C1}: \sum_{j=1}^{J} u_{ij} = 1, \quad \forall i \in \mathcal{I}, \\
& \mathrm{C2}: n_{i} \leq Q_{i}, \quad \forall j \in \mathcal{J}, \\
& \mathrm{C3}: u_{ij} \in \{0,1\}, \quad \forall i, j.
\end{aligned}
\end{equation}
where $u_{ij} = 1$ if AV $i$ is assigned to BS $j$, and zero otherwise \cite{yan2022gc}. The outputs can be decoded as a feature matrix representing the BS-User assignment by:

\begin{equation}
\label{eq:bs-user-matrix}
{u} \triangleq 
\begin{pmatrix}
{u}_{1,1} & \cdots & {u}_{1,N_{\mathcal{J}}} \\
\vdots & \ddots & \vdots \\
{u}_{N_{\mathcal{I}},1} & \cdots & {u}_{N_{\mathcal{I}},N_{\mathcal{J}}} \\
\end{pmatrix}
\subset \{0, 1\}^{N_{\mathcal{I}} \times N_\mathcal{J}}
\end{equation}

\section{Quadratic Unconstraint Binary Optimization}
\label{sec:qubo}

In this study, we aim to optimize the BS-User association matrix. 
Since $u_{ij}$ in (\ref{eq:optimization_problem}) represents a binary variable, this binary nature aligns with the Quadratic Unconstraint Binary Optimization (QUBO) problem \cite{kochenberger2014unconstrained}. Both the objective function and constraints in (\ref{eq:optimization_problem}) can be expressed as quadratic functions of binary variables.
Since conventional quantum optimization methods, such as fully separable variational states \cite{diez2021quantum}, impair quantum encoding performance and affect training efficiency, we apply QUBO to reformulate the problem outlined in (\ref{eq:optimization_problem}).

This section will introduce QUBO, a framework closely related to our problem formulation in equation (\ref{eq:bs-user-matrix}). 
Assuming a cost function $f(x)$ includes products of boolean variable $x_i \in \{ 0,1\}$. An matrix $Q \in \mathbb{R}^{n \times n}$ is $N$-by-$N$ square symmetric matrix of coefficients, we would like to minimize overall system cost function as \cite{diez2021quantum},
\begin{equation}
\label{eq:system_cost}
    E(\Vec{\mathbf{x}}) = \sum_{ij} x_i Q_{ij}  x_j,
\end{equation}

where $\Vec{\mathbf{x}}$ is an binary $N$-dimensional vector. $\mathbf{x}$ is a vector representing the binary assignment vector for BS $i$. Note that
$x_i x_i = x_i$ for binary variables.

Considering the system model presented in Section \ref{sec:system_model}, a QUBO problem can be mapped to an Ising spin model \cite{barkoutsos2020improving}. Therefore, the optimization of the system cost function in (\ref{eq:system_cost}) can be derived to find minimum energy state of an Ising Hamiltonian for an $N$-qubit system. Each qubit represents the vehicle-BS association. We can apply the change of variable $x_i \rightarrow \frac{1}{2} (1 + \sigma_i^z)$, where $\sigma_i^z$ is the Pauli-$Z$ matrix acting on the $i$-th qubit.
We convert this expression to an Ising Hamiltonian form by changing variables from binary variables $ x_i  $ to spin variables $ \sigma_i^z $ as

\resizebox{0.95\linewidth}{!}{
  \begin{minipage}{\linewidth}
  \begin{align*}
    E(\Vec{\mathbf{x}}) = & \sum_{ij} Q_{ij} \left( \frac{1}{2}(1 + \sigma_i^z) \right) \left( \frac{1}{2}(1 + \sigma_j^z) \right) \\
    = & \; \frac{1}{4} \bigg( \sum_{ij} Q_{ij} + \sum_{ij} Q_{ij} \sigma_i^z + \sum_{ij} Q_{ij} \sigma_j^z + \sum_{ij} Q_{ij} \sigma_i^z \sigma_j^z \bigg) \\
    = & \; \frac{1}{4} \left( \sum_{ij} Q_{ij} + \sum_i Q_{ii}\right) + \frac{1}{2} \sum_{i} \left( \sum_{j} Q_{ij} \right) \sigma_i^z + \frac{1}{2} \sum_{i<j} Q_{ij} \sigma_i^z \sigma_j^z \\
    = & \;  \frac{1}{4} \sum_{ij} Q_{ij} - \frac{1}{2} H .
\end{align*}
  \end{minipage}
}

where $ H $ is the Ising Hamiltonian \cite{barkoutsos2020improving}. 
Since $Q_{ij}$ contains both positive and negative coefficients, so the optimization is challenging.
The equation (\ref{eq:system_cost}) can be rewritten as

\begin{equation}
\label{eq:Hamiltonian}
        f(\sigma_i^z) =  \frac{1}{4} \sum_{ij}Q_{ij}  -\frac{1}{2}H ,
\end{equation}
where 
\begin{equation}
        H = \sum_{i < j} -Q_{ij} \sigma_i^z \sigma_j^z - \sum_i \left( \sum_j Q_{ij} \right) \sigma_i^z,
\end{equation}

Regarding equation \eqref{eq:optimization_problem} and \cite{kochenberger2014unconstrained}, the final QUBO problem can be formulated as:
\begin{equation}
\begin{aligned}
H_{\text{QUBO}} = & - \sum_{i=1}^{I} \sum_{j=1}^{J} \text{WR}_{ij} \, u_{ij} + 
 \lambda_1 \sum_{i=1}^{I} \left( \sum_{j=1}^{J} u_{ij} - 1 \right)^2 + \\
& \lambda_2 \sum_{i=1}^{I} \left( \sum_{j=1}^{J} u_{ij} - Q_i \right)_+^2
\end{aligned}
\end{equation}
where $ (x)_+ = \max(0, x) $, ensuring the penalty applies only when $n_i$ exceeds $Q_i$. $ \lambda_1 $ and $ \lambda_2 $ are the balancing factor on C1 and C2 in \eqref{eq:optimization_problem} .

\section{Proposed CVaR based VQE Approach}
 This section introduces our proposed CVaR-VQE method, designed to address the QUBO problem presented in the previous section. Compared to existing approaches, the CVaR-VQE method achieves accurate results with fewer samples and offers enhanced management of decoherence and operational errors. 
Specifically, this section first introduces quantum neural networks (QNNs), which serve as the practical architecture for executing quantum algorithms with parameterized gates. Then, we present variational quantum eigensolver (VQE), which is a key application of variational quantum circuits and is specifically designed to find the ground state energies of quantum systems. 
VQE combines quantum and classical computing in a hybrid approach, where the quantum circuit performs state preparation and measurement, while classical optimization adjusts the circuit parameters. As an enhancement to standard VQE, we further introduce the Conditional Value at Risk-Variational Quantum Eigensolver (CVaR-VQE) to focus on outlier examples to improve the training and evaluation efficiency under uncertainty.

\label{sec:CVaR_VQE}

\subsection{Parametrized Quantum Operations and Variational Quantum Algorithms} 
\label{subsection:parametrized}

\begin{figure}[!htb]
\centering
\begin{tikzpicture}[thick,scale=0.45, every node/.style={scale=0.7}]

\path (-14.2,-2.75) node[anchor=base, fill=white, align=center] {\normalsize $\Vec{\mathbf{x}}$};

\path (-13,-2.5) node[anchor=base, fill=white, align=center, rotate=90] {\scriptsize INPUT}; 
\draw[->,thick] (-13.3,-3.6)--(-12.9,-3.6);
\draw[->,thick] (-13.3,-1.4)--(-12.9,-1.4);

\path (0,0) node[anchor=base, align=center, minimum size=1cm](base){ }; 
\path (-12,-1) node[minimum size=0.8cm] (ket1) {\scriptsize $\ket{0}\,\,\,$}; 
\path (-12,-2) node[minimum size=0.8cm] (ket2) {\scriptsize $\ket{0}\,\,\,$};
\path (-12,-3) node[minimum size=0.8cm] (dots) {\scriptsize \vdots};
\path (-12,-4) node[minimum size=0.8cm] (ketn) {\scriptsize $\ket{0}\,\,\,$};

\draw[-] (-11.8,-1)--(-11,-1);
\draw[-] (-11.8,-2)--(-11,-2);
\draw[-] (-11.8,-3)--(-11,-3);
\draw[-] (-11.8,-4)--(-11,-4);

\path (-9,-2.6) node[draw,thick,anchor=base, align=center, minimum width=2.5cm, minimum height=2.5cm] (U1) {\small $\mathcal{U}_x \ket{0}^{\otimes S} = \ket{\psi_x}$}; 
\path (-9,-5.8) node[anchor=base, align=center] (label1){\scriptsize Feature mapping};
\path (-9,-6.4) node[anchor=base, align=center] (label1){\scriptsize $\mathcal{U}_x$};

\draw[-] (-7,-1)--(-6,-1);
\draw[-] (-7,-2)--(-6,-2);
\draw[-] (-7,-3)--(-6,-3);
\draw[-] (-7,-4)--(-6,-4);

\path (-3.7,-2.6) node[draw,thick,anchor=base, align=center, minimum width=3cm, minimum height=2.5cm] (G1) {\small $\mathcal{G}_\theta \ket{\psi_x} = \ket{g_\theta(x)}$}; 
\path (-3.7,-5.8) node[anchor=base, align=center] (label1){\scriptsize Variational Model};
\path (-3.7,-6.4) node[anchor=base, align=center] (label1){\scriptsize $\mathcal{G}_\theta$};

\draw[-] (-1.4,-1)--(-0.5,-1);
\draw[-] (-1.4,-2)--(-0.5,-2);
\draw[-] (-1.4,-3)--(-0.5,-3);
\draw[-] (-1.4,-4)--(-0.5,-4);

\node[meter](meter) at (0.2,-1) {};
\node[meter](meter) at (0.2,-2) {};
\node[meter](meter) at (0.2,-3) {};
\node[meter](meter) at (0.2,-4) {};

\path (1.5,-1) node[minimum size=0.8cm] (out1) {\scriptsize $z_1$};
\path (1.5,-2) node[minimum size=0.8cm] (out2) {\scriptsize $z_2$};
\path (1.5,-3) node[minimum size=0.8cm] (dots) {\scriptsize \vdots};
\path (1.5,-4) node[minimum size=0.8cm] (outn) {\scriptsize $z_S$};

\path (2.8,-2.5) node[anchor=base, fill=white, align=center, rotate=90] {\scriptsize BINARY OUTPUT};
\draw[->,thick] (2.5,-4.5)--(2.9,-4.5);
\draw[->,thick] (2.5,-0.5)--(2.9,-0.5);

\path (1,-5.8) node[anchor=base, align=center] (label1){\scriptsize measurement};
\path (1,-6.4) node[anchor=base, align=center] (label1){\scriptsize $f(z) = u$};
\path (3.4,-2.75) node[anchor=base, fill=white, align=center] {\normalsize $u$};
\draw[dotted] (-0.8,-0.1) rectangle (2.2,-4.8); 
\end{tikzpicture}
\caption{Overview of the quantum neural network (QNN)} 
\label{fig_qnn_model_intro}
\end{figure}
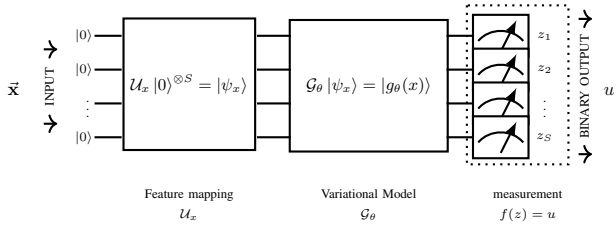

The basic structure of a quantum neural network (QNN) used in this study is illustrated in Fig.~\ref{fig_qnn_model_intro} \cite{abbas2021power}. 
The system initiates by transforming the input vector $ \Vec{\mathbf{x}} \in \mathbb{R}^{s_{\mathrm{input}}} $ into an $ S $-qubit quantum state within the Hilbert space through the application of a feature mapping operation $ \ket{\psi_x} := \mathcal{U}_x \ket{0}^{\otimes S} $. Subsequently, this quantum state undergoes evolution via a parameterized variational circuit, denoted as $ \ket{g_{\theta}(x)} := \mathcal{G}_{\theta} \ket{\psi_x} $, wherein the parameters $ \theta \in \Theta $ are optimized to minimize a specified loss function. The process culminates in a quantum measurement, yielding a classical bit string $ z = (z_1, \ldots, z_S) $, which is then processed through a classical function $ f $ to produce the final output BS-user assignment $ u := f(z) $.

This approach can be described through a general parameterized quantum operation, where the quantum state $ |\Psi(\vec{\theta}) \rangle $ is produced by sequentially applying layers of parameterized quantum gates:
\begin{equation}
    |\Psi(\vec{\theta}) \rangle = \mathcal{U}_L(\theta_L) \dots \mathcal{U}_1(\theta_1) \ket{\Psi_0} 
\end{equation}

\begin{figure}[!htb]
    \centering
    \resizebox{1\linewidth}{!}{
        \begin{quantikz}[row sep=0.5cm, column sep=0.3cm]
            \lstick{$\ket{0}$} & \gate{Y(\theta_1)} & \ctrl{1} &  \qw     & \ctrl{1} & \gate{Y(\theta_4)} & \ctrl{1} &  \qw     & \ctrl{1} & \gate{Y(\theta_7)} & \meter{0/1} \\
            \lstick{$\ket{0}$} & \gate{Y(\theta_2)} & \gate{Z} & \ctrl{1} & \octrl{1} & \gate{Y(\theta_5)} & \gate{Z} & \ctrl{1} & \octrl{1} & \gate{Y(\theta_8)} & \meter{0/1}  \\
            \lstick{$\ket{0}$} & \gate{Y(\theta_3)} &  \qw     & \gate{Z} & \gate{Z} & \gate{Y(\theta_6)} &   \qw     & \gate{Z} & \gate{Z} & \gate{Y(\theta_9)} & \meter{0/1} \\
        \begin{tikzpicture}[overlay]
            \draw[black, dashed, rounded corners]
                ($(0.2cm,0.4cm)$) rectangle ($(4.1cm,4.7cm)$);
        \end{tikzpicture}
        \begin{tikzpicture}[overlay]
            \draw[black, dashed, rounded corners]
                ($(4.2cm,0.4cm)$) rectangle ($(8.1cm,4.7cm)$);
        \end{tikzpicture}
        \end{quantikz}
    }
    \caption{Overview of the Variation Quantum Circuit (Ansatz)}
    \label{fig:vqe_varform}
\end{figure}
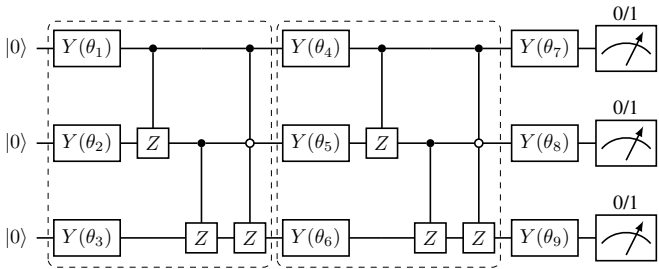

Fig.~\ref{fig:vqe_varform} further illustrates such a variational form $\mathcal{G}_{\theta}$.
Notice that this variational form spans all basis states. The input qubits are initialized to $\ket{0}$, undergo rotations with $Y(\theta)$ gates, and interact with each other via controlled-$Z$ gates, creating an entangled state prepared for quantum processing.  Fig.~\ref{fig:vqe_varform} shows a typical variational form for $3$ qubits depth $p=2$, where the circuit depth $p$ represents the number of layers in the quantum circuit, with each layer consisting rotation operations and entangling controlled-$Z$ gates applied to the qubits. 
In Fig.~\ref{fig:vqe_varform}, the dashed rectangle frames a variational quantum circuit.

\subsection{Variational Quantum Eigensolver Algorithm}
Variational quantum eigensolver (VQE) aims to find the minimum eigenvalue $\lambda_{\mathrm{min}}$ and corresponding Hamiltonian $H$ to solve the corresponding QUBO problem \cite{diez2021quantum} in terms of parametrized quantum operations in Section \ref{subsection:parametrized}.

\begin{equation}
\label{eq:vqe_qubo}
    \lambda_{\mathrm{min}} \le E(\vec{\theta})  = \langle \Psi(\vec{\theta}) | H | {\Psi(\vec{\theta})} \rangle,
\end{equation}
Thus, we can find the approximate optimal $\Vec{\theta}_{o}$ to approximate $E(\vec{\theta})$ to $\lambda_{\mathrm{min}}$ by  $\Vec{\theta}_{o} = \min_{\theta}E(\vec{\theta})$. VQE processes the wave function $|\Psi(\vec{\theta})\rangle$ in quantum machine, cost measured by energy cost $E(\vec\theta_{(k)})$ defined in (\ref{eq:system_cost}). Moreover, Quantum machine still relies on classical optimization approach to create an optimized sequence of parameters$\{\vec\theta_{(k)}\}$ to minimize $E(\vec{\theta})$.

VQE is an effective method for QUBO problems, where the energy function in \eqref{eq:Hamiltonian} consists of commuting observables. For a QUBO problem with $N$ vehicle-BS associations, the algorithm uses $N$ qubits initialized in a parameterized state $|\Psi(\vec\theta)\rangle$. The system energy is estimated by measuring all qubits in the Pauli-$z$ basis, producing values string $\{\mathbf{s}_i\in\{-1,1\}\}$ that convert to a bit string ${x_i}$ and corresponding QUBO cost $E_n$ by \eqref{eq:system_cost}. The energy expectation value is approximated through $K$ samples in \cite{diez2021quantum} as,
\begin{equation}
  \bar{E}_K(\vec\theta) := \frac{1}{K}\sum_{k=1}^{K}E_k \to E(\vec\theta),\,K\to\infty
  \label{samplemean}
\end{equation}

\subsection{Conditional Value at Risk-Variatiational Quantum Eigensolver (CVaR-VQE)}

The VQE algorithm faces a limitation in treating all samples with equal weight. The inclusion of outliers can significantly impair training efficiency and convergence. Denoting $P(E;\vec\theta)$ as the energy distribution associated with quantum states, we can reformulate the cost function of VQE from (\ref{eq:vqe_qubo}) to:
\begin{equation}
\begin{aligned}
\bar{E}_K(\vec\theta) &\simeq \mathbb{E}[E(\vec{\theta})] \\
&= \langle  \Psi(\vec{\theta})| H |\Psi(\vec{\theta}) \rangle \\
&= \int_{-\infty}^{\infty} P(E;\vec{\theta})E dE ,
\end{aligned}
\label{eq:old_vqe_loss}
\end{equation}

Authors in \cite{barkoutsos2020improving} proposed CVaR-VQE that modifies the cost function in (\ref{eq:old_vqe_loss}) to address the outlier issue. This approach focuses on the lowest $\alpha$ bit-strings of quantum samples, rather than utilizing all samples, by sorting and selecting them according to their energy values. Specifically, given sorted sampled energies $\{E_1< E_2< \ldots< E_K\}$ and taking a proportion $\alpha$ of the lowest energy outcomes, we can express:

\begin{equation}
    \textnormal{CVaR}_\alpha(E(\vec{\theta})) \approx \frac{1}{\lfloor\alpha K\rfloor}\sum_{k=1}^{\lfloor\alpha K\rfloor}E_k,
    \label{samplemeanCVaR}
\end{equation}
As the number of samples increases, the CVaR-VQE estimator \cite{diez2021quantum} converges to:\begin{equation}
\begin{split}
    \textnormal{CVaR}_\alpha(E(\vec{\theta})) &= \mathbb{E}[E(\vec{\theta})|E\leq F_{E}^{-1}(\alpha)] \\ & =\frac{1}{\alpha} \int_{-\infty}^{E_\alpha(\vec{\theta})} P(E;\vec{\theta})E dE\;\;,
\end{split}
\end{equation}
where $F_E$ denotes the cumulative density function of $E$, and $E_\alpha(\vec{\theta})$ satisfies:
\begin{equation}
    \int_{-\infty}^{E_\alpha(\vec{\theta})} P(E;\vec{\theta})dE = \alpha\,.
\end{equation}
The key insight behind this methodology is that optimizing the performance of the most promising samples (those with lowest energies) can be more effective than improving the overall average \cite{diez2021quantum}. This approach effectively concentrates the wave function's probability distribution toward the lowest-energy states, potentially leading to more efficient optimization.

\section{Case Study: BS-User Association in VNet}
\label{sec:case}
\begin{figure*}
    \centering
    \includegraphics[width=1\linewidth]{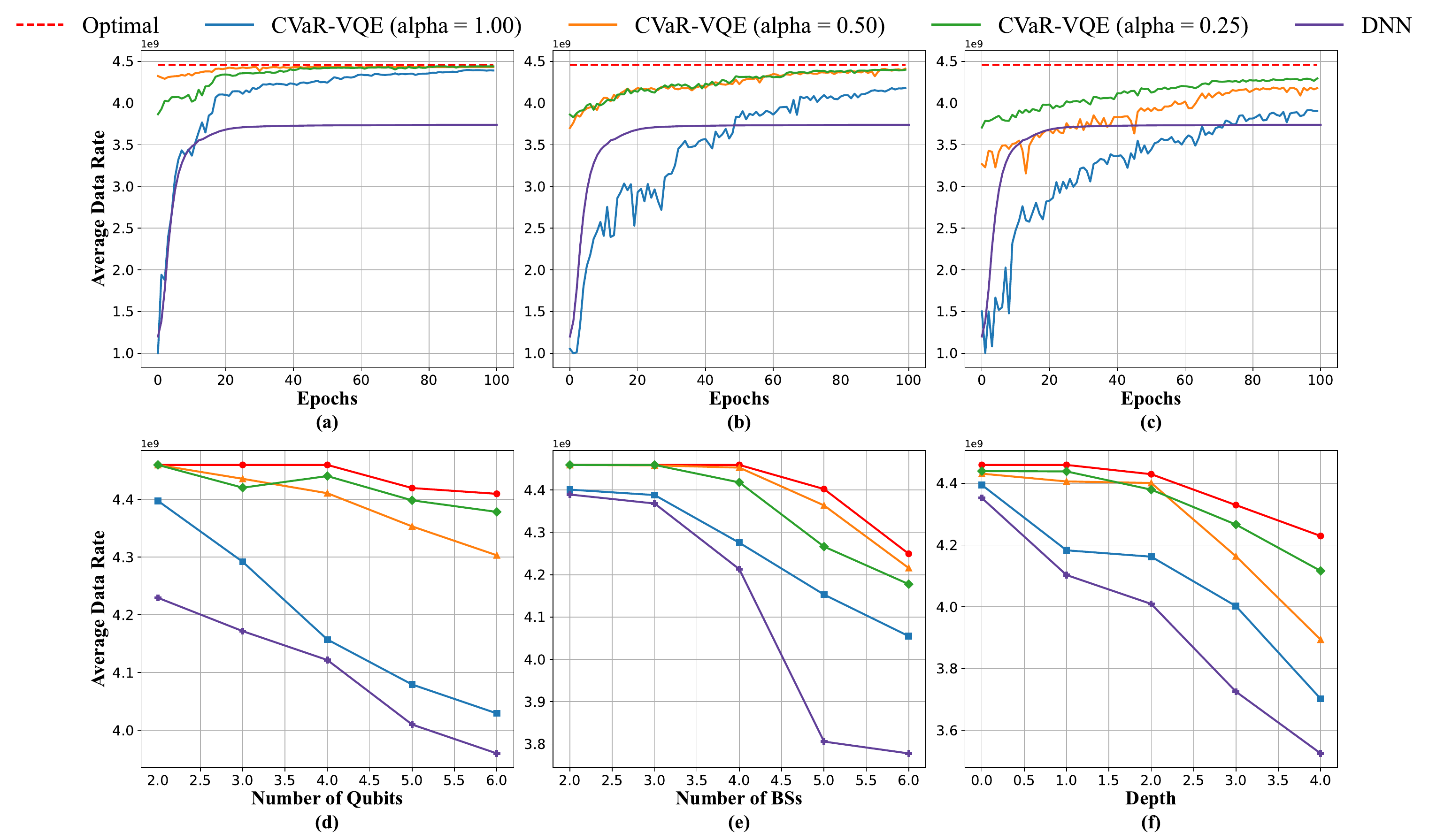}
    \centering
    \caption{Simulation results and comparisons}
    \label{fig:simulation}
\end{figure*}

\subsection{VNet Parameter Settings}

This section begins by outlining the simulation parameters and hyper-parameter configurations of the VNet and the proposed CVaR-VQE. Next, we provide a description of the benchmark DNN algorithm under consideration. Lastly, we present our primary results along with a detailed discussion on CVaR-VQE. The DNN model was trained and evaluated in Python using the PyTorch framework, while the CVaR-VQE model was trained and tested in Python using IBM's Qiskit library.

In our setup, AV users are assumed to be random distributed on the $1000$m single direction 4 lanes highway.
The molecular absorption coefficient, \( k_{a} \), is defined as 0.05 m\(^{-1}\) with a water vapor concentration of 1\%. This absorption coefficient is selected based on realistic database values, corresponding to a central frequency of 1.0 THz \cite{mobility}. For generalization, we normalize the sum rate by the transmission bandwidth. The RF transmission frequency is fixed at 2.1 GHz, with a path loss exponent \( \rho \) of 2.5. The antenna gains for THz transmission and reception, $G_{T}^{\text{tx}}$ and $G_{T}^{\text{rx}}$, are set to 25 dB, while the RF transmitter and receiver antenna gains are set to 0 dB. The transmit power for all base stations is standardized at 1 W, and the noise power is set at -70~dBm.

 \subsection{DNN Benchmark}
 In the benchmark scheme, the DNN generates a user association profile with continuous values, where each association is represented as \( 0 \leq u_{ij} \leq 1, \:\forall i \in \mathcal{I},\, \forall j \in \mathcal{J} \). The performance of the proposed CVaR-VQE approach is evaluated by comparison with this benchmark, where problem \(\mathcal{P}\) is solved to optimality.
DNN benchmark consists of a five-layer architecture designed specifically for the vehicle-BS association problem. The arhitecture includes an input layer accepting 16-dimensional flattened SINR matrices, 3 hidden layers with dimensions of 64, 128, and 256 neurons respectively, and a decoder layer reducing back to 64 neurons. The NN employs ReLU activation functions in the hidden layers, followed by a softmax operation on the output layer.

 Unless specified otherwise, the number of epochs, the batch size, and the learning rate for the benchmark DNN approach with 4 AVs and 4 base stations are set to \(100\), \(128\), and \(0.0001\), respectively.  The objective function is based on the summation data rate, as defined in  (\ref{eq:optimization_problem}).

\subsection{Proposed CVaR-VQE Framework}
{
The performance evaluation of our proposed CVaR-VQE framework was conducted through extensive experiments comparing it against conventional DNN and optimal approaches. Our analysis encompasses three key dimensions: \textbf{(1)} number of qubits, \textbf{(2)} number of base stations (BSs), and \textbf{(3)} circuit depth $p$. The experiments were conducted with different CVaR confidence levels ($\alpha \in \{0.25, 0.50, 1.00\}$).

\subsubsection{Convergence Analysis}
The training is based on 4 BSs and 4 AVs. Fig.~\ref{fig:simulation}(a-c) corresponds to different variation quantum circuit depth  $p = 1,2,3$, respectively.
We observe several key patterns in the convergence behavior:
The CVaR-VQE approach with $\alpha = 1.00$ demonstrates the fastest initial convergence, reaching near-optimal performance within the first 20 epochs. However, it shows some instability in the later epochs.
With $\alpha = 0.50$, we observe a more stable convergence pattern, though it takes approximately 40-50 epochs to reach comparable performance to the optimal solution.
The setting of $\alpha = 0.25$ exhibits the most conservative convergence behavior, with a slower but more stable approach toward the optimal value.
Overall, our CVaR-VQE framework achieves  $\mathcal{O}(\frac{1}{\alpha^2 K})$ variance based on $K$ samples input.

Fig.~\ref{fig:simulation}(a-c) indicates that results for $\alpha=0.25$ and $\alpha=0.5$ are practically similar.
However, taking $\alpha=1.00$, the probability of choosing an initial ground state increases to $5\%$ on average, after which it drops almost to zero even though the objective function continues to improve. 
The CVaR-VQE approach is primarily concerned with identifying the optimal performance-energy efficiency trade-off, with a particular emphasis on the lowest $\alpha$ examples.

\subsubsection{Evaluation based on different number of qubits}
As illustrated in Fig.~\ref{fig:simulation}(d), increasing the number of qubits from 2 to 6 leads to a consistent degradation in average data rate across all approaches.
The conventional DNN experiences the steepest decline in performance.
CVaR-VQE with \( \alpha = 0.25 \) maintains the closest proximity to the optimal solution.
The CVaR-VQE approach remains consistently high data rate across different qubit inputs. A greater number of qubits makes the quantum machine more complex, which reduces performance.

\subsubsection{Evaluation based on different number of BSs}
Fig.~\ref{fig:simulation}(e) demonstrates the impact of increasing the number of BSs. Performance degradation becomes more pronounced after 4 BSs since producing more handover switch penalty on connecting new BS. Additionally, the gap between optimal and CVaR-VQE solutions widens with an increased number of BSs. Increasing the number of BSs also increases the chance of handovers, as AVs are encouraged to connect to new, vacant, and available BSs.

\subsubsection{Evaluation based on VQE circuit depth \( p \)}
In this case, we add multiple layers of Convolutional Neural Networks in the DNN benchmark.
The effect of varying depth parameters (Fig.~\ref{fig:simulation}(f)) reveals that optimal performance is achieved at $p=1$. Performance degradation is observed if \( p>2 \), as excessive layers in variational quantum circuits impair training performance by increasing quantum entanglement and interaction among qubits.

\section{Conclusion}
\label{sec:conclusion}

Quantum computing holds transformative potential for next generation wireless networks. This study proposed a CVaR-VQE framework tailored to address user association challenges within vehicular networks. By leveraging a hybrid quantum-classical approach, our solution delivers a substantial performance improvement, surpassing conventional DNN methods by 23.5\%. This demonstrates the unique capacity of CVaR-enhanced VQE to handle the dynamic nature of vehicular networks and optimize BS-User assignment under uncertainty.
Future research could extend this framework to more varied vehicular network scenarios, integrating additional parameters like network load balancing and user mobility dynamics.

\bibliographystyle{IEEEtran}
\bibliography{main.bib}

\end{document}